# RFPPO：Motion Dynamic RRT based Fluid Field - PPO for Dynamic TF/TA Routing Planning

*Rongkun Xue, Jing Yang\*, Yuyang Jiang, Yiming Feng, Zi Yang, Member, IEEE*

*Abstract*— Existing local dynamic route planning algorithms, when directly applied to terrain following/terrain avoidance, or dynamic obstacle avoidance for large and medium-sized fixed-wing aircraft, fail to simultaneously meet the requirements of real-time performance, long-distance planning, and the dynamic constraints of large and medium-sized aircraft. To deal with this issue, this paper proposes the Motion Dynamic RRT based Fluid Field - PPO for Dynamic TF/TA Routing Planning. Firstly, the action and state spaces of the proximal policy gradient algorithm are redesigned using disturbance flow fields and artificial potential field algorithms, establishing an aircraft dynamics model, and designing a state transition process based on this model. Additionally, a reward function is designed to encourage strategies for obstacle avoidance, terrain following, terrain avoidance, and safe flight. Experimental results on real DEM data demonstrate that our algorithm can complete long-distance flight tasks through collision-free trajectory planning that complies with dynamic constraints, without the need for prior global planning.

## I. Introduction

In modern autonomous systems such as self-driving vehicles, and mobile robots, route planning algorithms are crucial, especially for drones or fighter jets in terms of terrain following and avoidance. From a decision-making perspective, route planning can be divided into two types: global static path planning and local dynamic path planning. Global static path planning aims to construct feasible routes on high-precision maps, while local dynamic path planning allows the system to perceive and adapt to real-time environmental changes. This includes responding to moving and static obstacles, and, integrating high-precision terrain maps, to generate or update optimal and safe paths in real time.

For large and medium-sized fixed-wing aircraft, due to the complexity of dynamic constraints and the vast scope of the maps, route planning algorithms typically employ both global static planning and local dynamic planning to complete flight missions. This approach requires prior offline global static planning to provide a global trajectory from the starting point to the destination, followed by local dynamic planning during flight based on environmental information to return to the global trajectory. However, when the prior map information is not accurate enough, or the actual flight environment is complex, the trajectory optimization performance of this method is limited. Furthermore, as large, and medium-sized fixed-wing aircraft have relatively weaker maneuverability while performing autonomous TF/TA, existing works are more focused on global static algorithms.

Global path planning primarily relies on designing heuristic and incremental functions. Koenig S et al. [1] proposed the LPA* algorithm, an incremental heuristic search variant of the A* algorithm. This algorithm categorizes nodes into three states: locally consistent, locally over-consistent, and locally under-consistent. It addresses the issue of changing edge costs on a finite grid map over time, as caused by changes in obstacles and grid points. This solves the efficiency problem of repeatedly using A* for research under these conditions. Moreover, in its planning process, the start and goal points are fixed, further limiting the power requirements of large and medium-sized fixed-wing aircraft. The Rapidly exploring Random Tree (RRT) algorithm aims to effectively search non-convex, high-dimensional spaces by randomly building a space-filling tree. This tree is incrementally constructed from randomly sampled points in the search space, with a bias towards exploring large unsearched areas, and can easily handle obstacles and differential constraints. Jiaming Fan and Xia Chen [2] proposed a drone trajectory planning based on a bidirectional APF-RRT* algorithm with target-biasing. This approach guides the generation of random sample points with a target-biasing strategy, uses a bidirectional RRT* algorithm to establish two alternating random search trees, and introduces an improved artificial potential field method in the bidirectional growth trees, significantly reducing the number of iterations.

Local dynamic programming algorithms focus more on real-time performance. Dave Ferguson [3] proposed the Field D* algorithm, which uses linear interpolation on grids to allow path points not to be limited to endpoints. This approach allows for smoother planned curves as changes in planning direction are no longer restricted to π/4 increments. However, this method does not consider dynamic factors, making it challenging to plan paths for large and medium-sized aircraft effectively. The DWA (Dynamic Window Approach) algorithm [4] samples multiple speed sets in the velocity space (v,w) and simulates their trajectories over a unit of time. It selects the optimal trajectory and corresponding (v,w) to drive the robot's movement. Unlike D*, DWA fully considers the robot's dynamics but only simulates and evaluates the next step, making it inefficient for avoiding obstacles like 'C' shaped ones. When extended to three-dimensional space, its computational load increases drastically, unsuitable for real-time planning. The Artificial Potential Field (APF) algorithm [5-7], known for its simpler theory, often experiences significant path

This work was supported by the National Natural Science Foundation of China under Grant No. 62073257
*Corresponding author.

Rongkun Xue, Jing Yang, Yiming Feng, Zi Zhou are with the School of Automation Science and Engineering, Xi'an Jiaotong University, Xi'an, Shaanxi 710049, China (e-mail: xuerongkun@stu.xjtu.edu.cn; jasmine1976@xjtu.edu.cn;3123154001@stu.xjtu.edu.cn;Yangzi@mail.xjtu.edu.cn ).
Yuyang Jiang is with the School of Software Engineering, Xi'an Jiaotong University, Xi'an, Shaanxi 710049, China (e-mail: Jiang_yuyang@stu.xjtu.edu.cn).

fluctuations when navigating through narrow spaces. Its performance is also hindered by the relationship between the artificial field and obstacle distribution, particularly near the target with multiple moving obstacles. The lack of information on obstacle movement complicates the design of a universal potential function, frequently leading to local optima issues for the agent. Virtual mechanics [8] treat agents and obstacles as particles. Due to vague definitions of attraction and repulsion and neglect of obstacle shapes, it struggles with effective obstacle avoidance, especially with local obstacles in real-world scenarios, often requiring pilot involvement in planning. Wang et al. [9] proposed an algorithm known as the Interfered Fluid Dynamical System (IFDS), drawing inspiration from the macroscopic characteristics of water flow in nature, which moves in a straight line in the absence of obstacles and smoothly navigates around them when encountered. This algorithm achieves planning curves superior to those of traditional potential field methods. Building upon the IFDS, Wu et al. [10] introduced a Model Interfered Fluid Dynamical System (MIFDS) within the MAF obstacle avoidance framework. This framework extends the Interfered Fluid concept by incorporating a constrained motion model for unmanned aerial vehicles.

Currently, artificial intelligence methods like deep learning and reinforcement learning demonstrate impressive performance in route planning. Sutton R S et al. [11] have explicitly represented policies with their own function approximators, updating them through the gradient of expected returns with respect to policy parameters. They proved that any differentiable function approximation of policy iteration converges to a local optimum policy, pioneering policy-based work in the field of deep reinforcement learning. Schulman J [12] proposed a policy gradient method for reinforcement learning, where sampling data is obtained through interaction with the environment, and a "surrogate" objective function is optimized using stochastic gradient ascent. This method, Proximal Policy Optimization (PPO), offers more generality than Trust Region Policy Optimization (TRPO) and strikes a good balance between sample complexity and efficiency. Deep network-based reinforcement learning gains feedback from the environment during agent-environment interactions and updates its decision-making network weights based on this feedback, ensuring maximization of its own estimates. This approach ensures model generality and provides more solutions to complex obstacle environments, such as multi-obstacle spaces and dynamic local obstacles. Chen et al. [13] introduced a distributed deep reinforcement learning-based obstacle avoidance path planning algorithm, significantly reducing the time needed for agents to complete avoidance tasks and reach their destinations. Moldovan and Abbeel [14] suggested that an agent is 'safe' if it meets a pervasiveness requirement, meaning it can reach every state it visits from any other state it visits, allowing for reversible mistakes. The field of reinforcement learning has made some contributions to safety. Mirchev et al. [15-16] used Q-learning methods and tree-based ensemble methods as function approximators to achieve high-level lane change control in highway scenarios. Their method impressively reduced collision probabilities. However, this approach might only be applicable in dual-lane change environments since it considers only one lane change option in the environment at any time. Kulkarni et al. [17] proposed a method combining reinforcement learning with supervised learning. In this approach, the agent uses function approximation in reinforcement learning to generalize planned trajectories, enabling trajectory planning in various environments with strong generalization capabilities.

The Proximal Policy Gradient algorithm, in a Model Free context, achieves policy gradient updates through continuous exploration and learning, demonstrating strong generalizability. Motion Dynamics RRT [18] can plan highly similar, reachable trajectories for large and medium-sized aircraft in a very short time, considering their dynamics. Both Artificial Potential Field and Interfered Fluid Dynamical System, using field construction methods, have shown strong performance in local dynamic obstacle avoidance. To achieve more efficient training and execution, we have integrated these methods into our framework. This integration allows us to utilize global planning information during training, while avoiding reliance on global path planning during execution, thus achieving dynamic obstacle avoidance and long-distance TF/TA. Moreover, our flight trajectory design strictly considers the dynamic model of large and medium-sized aircraft. By employing the Key Point method to optimize accessible space, we have enhanced training efficiency.

## II. METHOD

For the task of planning for large and medium-sized fixed-wing aircraft in complex terrain, we propose a framework named Motion Dynamic RRT based Fluid Field - PPO (RF-PPO). This framework can achieve collision-free trajectory planning that adheres to dynamic constraints without relying on prior global planning, thereby effectively facilitating long-distance flight planning.

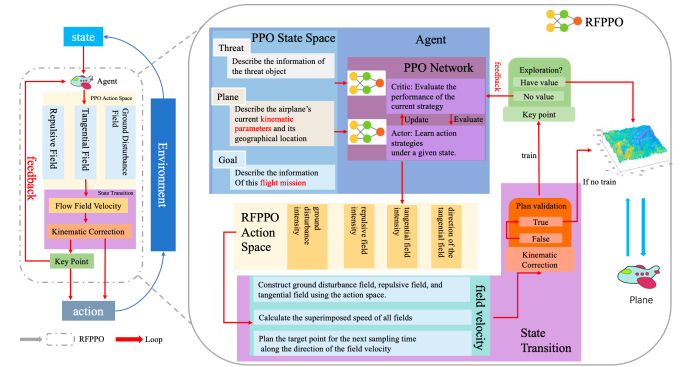

Fig. 1. Overview of our proposed RFPPO framework (*left*) and inner details of RFPPO pipeline (*right*)

Our framework's input state space includes information such as threats and the aircraft's own observations, while the action space covers the strength and direction for constructing flow fluid. By integrating field velocity with aircraft dynamics, we achieve more trackable trajectories during state transitions. In the Agent section, we introduce the state space and the RFPPO neural network. We then discuss the RFPPO action space based on field velocity and dynamic constraints, along with its state transition process. Additionally, this study combines the concept of safe reinforcement learning with the characteristics of TF/TA tasks, proposing a method for key point safety decision-making based on Motion Dynamics Rapidly exploring Random Trees (RRT). This not only achieves long-distance flight planning but also significantly reduces the algorithm's training time. Finally, our framework's effectiveness is successfully demonstrated through experiments on Digital Elevation Model (DEM) maps.

*A. Agent*

The state space and neural networks together constitute the intelligent agent we have defined. The task of this agent is to model the fluid field based on information from the state space, thereby deriving an efficient trajectory planning algorithm. The efficiency and simplicity of the fluid field ensure that it can accomplish flight missions within a smaller state space and a simple neural network, without the need for extensive environmental information.

*1) State space*

Our state space is composed of multiple positional vectors and informational scalars, which comprehensively detail the aircraft's current flight state and its obstacle avoidance challenges. It encompasses the vector distances from the aircraft's current position to both the start and target points, the vector length to the nearest point of any observable obstacles, the current altitude above ground, and flight state parameters, such as the heading and climb angles.

*2) RFPPO Network*

We employ the Proximal Policy Optimization algorithm for learning and decision-making, utilizing a parameterized probability distribution $\pi_\theta(a\mid s) = P(a\mid s;\theta)$ to replace the deterministic policy of the value function $\pi: s \rightarrow a$. Compared to other reinforcement learning algorithms, this class of algorithms holds unique advantages in handling continuous spaces.

We developed two fully connected neural network models, each with four identical layers (N = 4). ReLU (Rectified Linear Unit) activation functions are applied between these layers, while the final layer uses the 'tau' activation function for output range control. The actor network, aligning its input layer with the state space dimension, selects actions based on current state. Its output layer matches the action space dimension. And, the critic network assesses state values, concentrating on the long-term effects of the agent's present state on policy decisions.

Specifically，we make policy parameters $\theta_0$, and make clipping threshold $\epsilon$, and for k=0,1, 2,…, we collect set of partial trajectories $D_k$ on policy $\pi_k = \pi(\theta_k)$. So, estimate advantages $\hat{A}_t^{\pi_k}$ using any advantage estimation algorithm and we can take K steps of minibatch SGD. Subsequently, we update the actor network using $\mathcal{L}_{\theta_k}^{CLIP}(\theta) = \mathop{\mathbb{E}}_{\tau \sim \pi_k}\left[\sum_{t=0}^{T}\left[min(r_t(\theta)\hat{A}_t^{\pi_k}, \text{clip}(r_t(\theta), 1-\epsilon, 1+\epsilon)\hat{A}_t^{\pi_k})\right]\right]$ and train the critic network by calculating the difference between the discounted return $G_t = r_{t+1} + \gamma r_{t+2} + \cdots + \gamma^{T-t}r_{T+1} + \gamma^{T+1-t}v(s_{T+1})$ and the network output $Critic_{state}$ using the MSE loss function. The reward at time T, which is derived from the designed reward function based on the state space, will be discussed in section E. Additionally, γ represents the discount factor for rewards, used in reinforcement learning to balance the importance of immediate versus future rewards, thus guiding the agent to make rational decisions between long-term goals and short-term actions.

*B. RFPPO Action Space*

If the coordinates of the aircraft are directly defined as the action space in reinforcement learning, it necessitates the design of complex reward functions and neural network architectures to ensure the feasibility of the task and the compliance of the flight trajectory. Therefore, drawing inspiration from the concepts of Interfered Fluid Dynamical Systems (IFDS) and Artificial Potential Fields, we use four positive constants as parameters, transforming the action space from directly deciding the next coordinate point to responding to the perceived field of the current environment. Under this approach, flight path planning is conducted under the influence of this field.

- Ground Disturbance Intensity (β): The influence exerted by the ground on the agent is abstracted as an upward repulsive potential field emanating from the ground. When this value is large, the agent is more inclined to perform terrain-following tasks, thereby maintaining a greater distance from the ground. Conversely, when the value is small, the agent tends to engage in terrain avoidance, reducing its proximity to the ground.

- Repulsive Field Strength (ρ): This parameter represents the intensity of the repulsion exerted by an obstacle on the agent in that state. A higher value of ρ generates a stronger reaction force, pushing the agent away from the obstacle; conversely, a lower value results in a weaker force, potentially allowing the agent to approach the obstacle.

- Tangential Field Strength (σ): This parameter is used to quantify the strength of the flow field created by obstacles on the tangential plane. It not only generates a repulsive effect but also forms a guiding effect on the tangential plane. When this value is high, the agent's trajectory converges more rapidly towards the target point; conversely, when the value is low, the agent's trajectory becomes smoother, enhancing the transitional and continuous nature of the flight path.

- Tangential Field Angle ($\theta$): This parameter defines the direction of the tangential field, endowing the agent with the ability to find and follow flow field trajectories that conform to kinematic constraints within the flow field.

*C. State Transition*

Through the values in the action space, we can abstract the obstacles and environment into a fluid field, ultimately represented as the velocity of the field $\bar{u}$. By using $P_{t+1} = P_t + \bar{u} \cdot \Delta T$, we obtain the coordinates for the desired trajectory planning at the next moment, $P_{t+1(unrestricted)}$. Simultaneously, it's necessary to integrate adjustments based on the aircraft's dynamics. This correction ensures that while we follow the direction of the fluid field for state transitions, we also strictly adhere to the characteristics of the aircraft's dynamics. This results in the trajectory planning coordinates for the next moment, $P_{t+1(restricted)}$.

*1) Flow Field Velocity*

In this section, we elaborate on how to combine the Interfered Fluid Dynamical System and Artificial Potential Field to construct the fluid field, and ultimately derive the flow field velocity from the fluid field.

First, we quantify the threat of obstacles to the agent by (1).

$$F(\xi) = \left(\frac{x-x_0}{a}\right)^{2d} + \left(\frac{y-y_0}{b}\right)^{2e} + \left(\frac{z-z_0}{c}\right)^{2f} \quad (1)$$

In the Cartesian coordinate system, we typically set the geodetic origin of China in the DEM map as the origin. The spatial coordinates of the current agent are defined as $(x, y, z)$, and the centroid of the obstacle is set as $(x_0, y_0, z_0)$ with semi-axes lengths $(a, b, c)$. The values of d, e, and f further determine the shape of the obstacle. For example, setting d=1, e=1, f=1 results in an obstacle with a spherical shape.

*a) Repulsive Field*

Eq 2 quantifies the repulsion effect and direction of obstacles on the agent. When a large or medium-sized fixed-wing aircraft approaches an obstacle, the agent experiences repulsion in the direction of the normal vector to the surface of the obstacle.

$$\begin{cases} R_k(\xi) = -\dfrac{n_k n_k^T}{|F_k|^{\frac{1}{\rho_0}} \cdot n_k n_k^T} \\ n_k = \left[\dfrac{\partial F_k}{\partial x} \quad \dfrac{\partial F_k}{\partial y} \quad \dfrac{\partial F_k}{\partial z}\right]^T \end{cases} \quad (2)$$

Where, $F_k$ represents the threat generated by the obstacle, $\rho_0$ is one of the action spaces of the neural network, $d_0$ is the shortest Euclidean distance from the surface of the obstacle to the current agent, and $n_k$ is the normal vector of the obstacle's surface.

*b) Ground Disturbance Field*

To achieve TF/TA in complex terrains, we incorporate the concept of artificial potential fields. Through Eq 3, we define the ground disturbance field, which is always perpendicular to the ground surface.

$$M_\chi = I \cdot \beta_0 \cdot \ln\left(\frac{height_{agent}}{height_{safe}} + 1\right) \quad 3$$

Where, I is a three-dimensional matrix, $height_{safe}$ represents the minimum height above ground allowed for the agent, $height_{agent}$ denotes the actual height of the agent, and $\beta_0$ is one of the action spaces of the neural network.

*b) Tangential Field*

Eq 4 quantifies the strength and direction of the tangential guidance exerted by obstacles on the fluid.

$$T_k(\xi) = \frac{t_k n_k^T}{|F_k|^{\frac{1}{\sigma_k}} \|t_k\| \|n_k\|} \quad 4'$$

$\sigma_k$ is one of the action spaces of the neural network, and $t_k$ is computed using Eq 5. This value determines the direction of the field.

$$\begin{cases} t_{k,1} = \left[\dfrac{\partial F_k}{\partial y} \quad -\dfrac{\partial F_k}{\partial x} \quad 0\right]^T \\ t_{k,2} = \left[\dfrac{\partial F_k}{\partial x} \cdot \dfrac{\partial F_k}{\partial z} \quad \dfrac{\partial F_k}{\partial y} \cdot \dfrac{\partial F_k}{\partial z} \quad -\left(\dfrac{\partial F_k}{\partial x}\right)^2 - \left(\dfrac{\partial F_k}{\partial y}\right)^2\right]^T \end{cases} \quad 5$$

$$t_k'(P) = [\cos\theta_0 \quad \sin\theta_0 \quad 0]^T$$
$$t_k = \Omega_T^I t_k'$$

Where $t_{k,1}$ and $t_{k,2}$ are the basis vectors of the tangential reference frame, $\theta_0$ is one of the action spaces, and $\Omega_T^I$ represents the coordinate transformation matrix from the tangential reference frame to the inertial frame.

We quantify the fuel field using Eq 6. $M_k$ represents the influence of the kth obstacle on the field. In practical engineering applications, a constant $R_{conf}$ is established, meaning only disturbances from obstacles within a distance less than $R_{conf}$ from the agent are considered, thus reducing the data volume in the algorithm's iterative process.

$$M_K(\xi) = R(\xi) + T(\xi) + M_\chi$$
$$\bar{M} = \sum_{k=1}^{K} \omega_k \mathbf{M}_k$$
$$\omega_k = \begin{cases} 1 & K = 1 \\ \prod_{i=1, i \neq k}^{K} \dfrac{(F_i - 1)}{(F_i - 1) + (F_k - 1)} & K \neq 1 \end{cases} \quad 6$$

Finally, the velocity of the fuel field is obtained using Eq 7.

$$\bar{u} = \bar{M}\left(\mathbf{u} - \sum_{k=1}^{K} \omega_k \exp\left(\frac{-(F_k - 1)}{\lambda_k}\right) \mathbf{v}_k\right) + \sum_{k=1}^{K} \omega_k \exp\left(\frac{-(F_k - 1)}{\lambda_k}\right) \mathbf{v}_k \quad 7$$

Where $u$ represents the velocity of the initial flow field, which is considered as the cruising speed of the aircraft in practical missions. $v_k$ is the velocity of the kth obstacle, and $\lambda_k$ is a constant that characterizes the impact of the obstacle's velocity on trajectory planning.

*2) Kinematic Correction*

As mentioned in the state transition section above, we need to strictly consider the aircraft's dynamic constraints, adjusting $P_{t+1(unrestricted)}$ to $P_{t+1(restricted)}$ in both horizontal and vertical directions.

The aircraft's dynamic transition in the horizontal direction follows Eq 8.

$$\begin{bmatrix} x_2 \\ y_2 \\ \varphi_2 \end{bmatrix} = \begin{bmatrix} \cos\varphi_1 & -\sin\varphi_1 & 0 \\ \sin\varphi_1 & \cos\varphi_1 & 0 \\ 0 & 0 & 1 \end{bmatrix} \cdot \begin{bmatrix} \dfrac{1}{\rho_T} \cdot \sin\Delta\varphi \\ \dfrac{1}{\rho_T} \cdot (1 - \cos\Delta\varphi) \\ \rho_T \cdot v \cdot \Delta t \end{bmatrix} + \begin{bmatrix} x_1 \\ y_1 \\ \varphi_1 \end{bmatrix} \quad 8$$

we use a body coordinate system relative to the aircraft's tethered plumb line in a fixed ground coordinate system. The next state is obtained by rotating the parent node, which has a track azimuth angle of $\varphi_1$, around the axis by $\Delta\varphi$, where $\Delta\varphi = \rho_H v \Delta t$. The turning radius is $R_H = \frac{1}{\rho_H}$. Additionally, $\rho_H = \frac{g}{v^2} \tan\emptyset$, where $\emptyset$ is the roll angle.

In the vertical direction, the dynamic transition follows Eq 9. $(x, y, z)$ represents the position in the agent's coordinate system, while $\gamma$ and $\chi$ respectively denote the agent's climb angle and track heading angle. They represent the current velocity of the agent.

$$\begin{cases} \dot{x} = V\cos\gamma\cos\chi \\ \dot{y} = V\cos\gamma\sin\chi \\ \dot{z} = V\sin\gamma \\ \dot{V} = (n_x - \sin\gamma)g \\ \dot{\gamma} = g(n_z - \cos\gamma)/V \\ \dot{\chi} = n_y g / V\cos\gamma \end{cases} \quad 9$$

Therefore, for the current node p, the unrestricted next node $P_{t+1(unrestricted)}$, and the previous node $p_{pre}$, we calculate the track angle and roll angle in the horizontal direction, and the climb angle in the vertical direction through the inverse processes of Eq 8 and 9. These are then adjusted within the permissible range of the agent. Subsequently, a forward calculation is performed along the direction of the field's velocity, ultimately yielding an easy-to-follow and dynamically compliant $P_{t+1(restricted)}$.

*D. Key Decision Points in Motion Dynamics RRT*

*a) Motion Dynamics RRT*

The Motion Dynamics RRT [18] algorithm largely adopts the framework of the RRT* algorithm. However, to address the issue of low search efficiency and slow convergence in vast three-dimensional planning spaces, it introduces an innovative method: generating an elliptical constrained search area based on the current shortest path. This approach enables global path planning within a 500KM range in just 8 seconds, allowing for rapid rewards acquisition from key points during training.

In the choice between autonomous terrain following and terrain avoidance tasks, the cost function considers not only flight safety factors but also horizontal and terrain-following route costs. By incorporating the aircraft's dynamic performance constraints from RFPPO, the algorithm sets a weighted relationship between these two costs, allowing it to adaptively adjust the priority of terrain following and avoidance tasks based on real-world conditions. Additionally, to address the high randomness in result generation of the RRT* algorithm, Motion Dynamics RRT adopts a state transition based on motion dynamics equations when generating new nodes. This approach, while maintaining a certain randomness in sampling points, significantly reduces the randomness of effective new nodes. This ensures that the information obtained at the same key points across different batches is stable, preventing the high randomness of the RRT* algorithm from disrupting the convergence of our network training.

---

**Algorithm 1: MD-RRT***

**Input:** start_node $p_{start}$, goal_node $p_{goal}$
**Output:**
1:     Initialization $V \leftarrow \{p_{\text{start}}\}; E \leftarrow \emptyset; X_{\text{soln}} \leftarrow \emptyset;$
2:     **for** each $iter \in [1, \text{iter}_{\text{Max}}]$ **do**
3:         $c_{\text{best}} \leftarrow \min_{p_{\text{soln}} \in X_{\text{soln}}} \{\text{Cost}(p_{\text{soln}})\};$
4:         $p_{\text{rand}} \leftarrow \text{SampleInEllipse}(p_{\text{start}}, p_{\text{goal}}, c_{\text{best}});$
5:         $p_{\text{near}} \leftarrow \text{Nearest}(V, p_{\text{rand}})$    $p_{\text{new}} \leftarrow \text{Steer}(p_{\text{near}}, p_{\text{rand}})$
6:         **if** Collisionfree $(p_{\text{near}}, p_{\text{new}}, X_{\text{obs}})$ **then**
7:           $X_{\text{near}} \leftarrow \text{Near}(V, p_{\text{new}}, R_{\text{near}});$
8:           $p_{\text{father}} \leftarrow \text{ChooseParent}(X_{\text{near}}, p_{\text{new}}, p_{\text{near}});$
9:           $p_{\text{near}} \leftarrow p_{\text{father}}; V \leftarrow V \cup \{p_{\text{new}}\}; E \leftarrow E \cup \{(p_{\text{near}}, p_{\text{new}})\}$
10:        $G \leftarrow \text{Rewire}(G, p_{\text{new}}, p_{\text{near}})$
11:       **end if**
12:       **if** InGoalRegion $(p_{\text{new}})$ **then**
13:         $X_{\text{soln}} \leftarrow X_{\text{soln}} \cup \{p_{\text{new}}\};$
14:       **end if**
15:    **end for**
16:    Return $G = (V, E)$

---

*b) Ground Disturbance Field*

In the RFPPO training process, key decision points, $key_{list}$, are determined based on distance. Particularly, more decision points are set when the distance to the endpoint increases, to ensure effective exploration. At these points, we employ Motion Dynamics RRT sampling with the same time sampling rate as PPO for global planning. The RRT quickly ascertains whether the current state can reach the endpoint or identifies the time step $T_1$ when it cannot. If it's the latter, we terminate the training and return $R_{rrt} = T_1$. After implementing this algorithm, training on a relatively complex map only requires about 54 hours on a 3080, enabling our model to achieve real-time path planning within a 200KM range. This approach nearly saves 30% of the time compared to methods not using it, leading us to integrate this improvement into our framework.

---

**Algorithm 2: Key Point of RFPPO**

**Input:** $p_{state}, key_{list}, p_{goal}$
**Output:**
17:    Dis $=(p_{goal}(0) - p_{state}(0))^2 + (p_{goal}(1) - p_{state}(1))^2 + (p_{goal}(2) - p_{state}(2))^2$
18:    **if** Dis in the $key_{list}$ **then**
19:       $res, res_{numer}$ = Motion Dynamic RRT $(p_{state}, p_{goal})$
20:       **if** res == **True continue**
21:       **else** return reward=$res_{numer}$  done=**True break**
22:    **else continue**

---

*E. Reward Function*

The purpose of the reward function is to motivate the agent to achieve its objectives. Based on the characteristics of the TF/TA tasks, we designed the reward function according to Eq 10. The ratio $w_h/w_o$ influences the large aircraft's propensity for terrain avoidance and following, while $w_p$ and $w_r$ affect the aircraft's flight posture.

$$R_t = w_h r_h + w_o r_{obs} + w_p r_p + w_r r_{rrt} \qquad 10$$

For the first part $r_h$, we use Eq 13 to ensure that large aircraft can follow the terrain while maintaining the safest possible distance close to the ground.

$$r_h = -\chi\left(\frac{h_{down} - h}{h}\right) - \delta\left(\frac{h - h_{up}}{h}\right) - \frac{d_{now}}{d_{all}} \qquad 11$$

The weights of $\chi/\delta$ determine whether the flight trajectory is smoother or provides better concealment for the aircraft. $h$ represents the altitude above ground, with $h_{down}$ and $h_{up}$ being the minimum and maximum allowed altitudes, respectively. $d_{now}$ and $d_{all}$ respectively represent the agent's current distance to the target point and the total distance over the entire path planning process.

For the second part, $r_{obs}$, we use the method of Eq 12, employing a gradient-based reward approach to ensure that large aircraft stay as far away as possible from dynamic obstacles and choose safer penetration routes.

$$r_{obs} = -\alpha\left(\frac{d - R_{obs}}{R_{obs}}\right) - \beta\left(\frac{d - R_{obs} - R_{threaten}}{R_{obs} + R_{threaten}}\right) \qquad 12$$

Where $d$ represents the current closest distance between the agent and an obstacle, while $R_{obs}$ and $R_{threaten}$ denote the minimum and maximum allowed distances from the center of an obstacle, which can be considered as the radar's fire control range and warning range, respectively.

The third part, $r_p$, is usually 0 and only imposes limits through Eq 15 when the deviation angle and climb angle exceed $\varphi_{good}$. This ensures that the paths planned by the controller are easy to follow. The purpose of this sparse reward is to encourage the agent to focus more on TF/TA.

$$r_p = -k \ln \frac{|\varphi_{climb} - \varphi_{good}|}{\varphi_{good}} - \phi \ln \frac{|\varphi_{track} - \varphi_{good}|}{\varphi_{good}} \qquad 13$$

Where $\varphi_{climb}$ and $\varphi_{track}$ are the agent's climb angle and track angle, respectively, while $\varphi_{good}$ is the maximum angle allowed by the dynamics. The values of $\kappa, \varphi$ are generally around 1.

In the fourth part, $r_{rrt}$ is typically 0, only changing when the agent is at a key point. Here, we use Motion Dynamics RRT to assess if the agent can reach its destination while meeting dynamic constraints. If deemed unreachable, $r_{rrt}$ is set to -T, where T represents the predicted number of future failed steps.

III. EXPERIMENTS

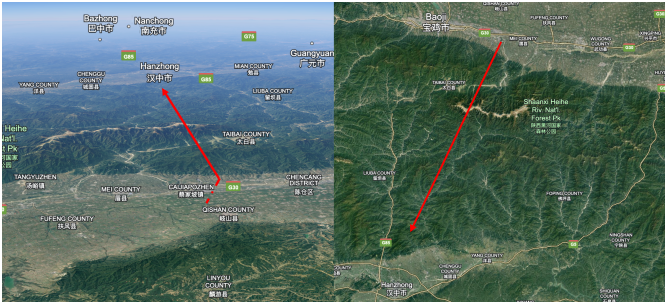

Fig. 2. Experimental scenario.

To validate the effectiveness and robustness of our algorithm, we select a diverse Digital Elevation Model (DEM) map as the basis for training and testing. This map is approximately 100KM*100KM in size, encompassing terrains such as canyons, forests, and plains. We randomly position 10 groups of dynamic threats across the map, featuring movement patterns including sine, circular, tangent, and straight lines. In the experiments, the takeoff point is randomly located within a 20km area around 34°23′55″N 107°25′31″E, and the destination is within a 20km area around 37°42′04″N 113°59′58″E. We randomly select 20 points, forming 10 sets of start and target points, and conduct multiple flight path plannings for each set to ensure the stability of the results. The simulations are run on a computer equipped with a CPU Intel Core i3-8100 3.60 GHz.

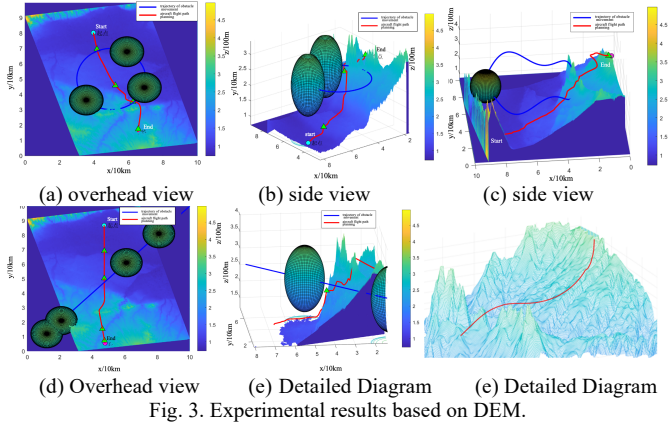

(a) overhead view  (b) side view  (c) side view

(d) Overhead view  (e) Detailed Diagram  (e) Detailed Diagram

Fig. 3. Experimental results based on DEM.

In the specific experiment, we set the number of training rounds to 10,000, with a batch size of 512. The learning rates for the critic and actor networks are both 0.001. The reward decay rate $\gamma'$ is 0.98. To ensure our simulation scenario is realistic, the agent can only obtain dynamic threat information when it is within 10KM of the threat. There is also a 0.05 probability of not receiving threat information. Fig. 3 (a) and Fig. 3 (b) show the overall top and side views of the threat object moving in a circular path, respectively. Fig. 3 (c) displays the overall side view when the threat object moves in a sine-cosine pattern. Fig. 3 (d), Fig. 3 (e), and Fig. 3 (f) show the overall top view of the threat object moving in a straight line, a detailed local view when the aircraft is closest to the threat, and another detailed local view when the aircraft is closest to the threat.

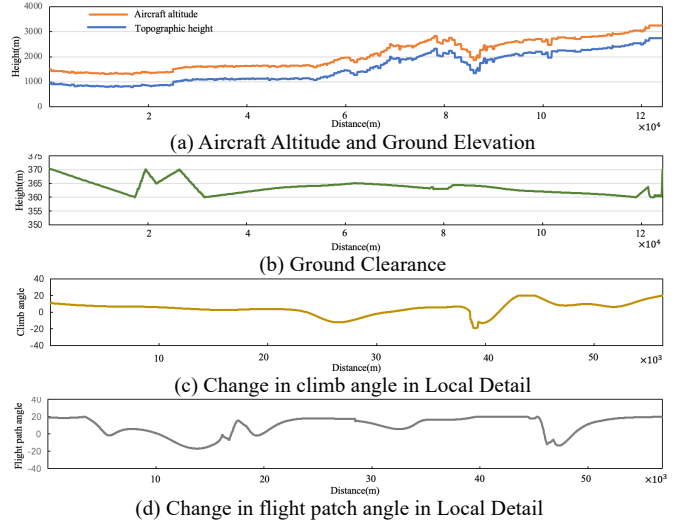

(a) Aircraft Altitude and Ground Elevation

(b) Ground Clearance

(c) Change in climb angle in Local Detail

(d) Change in flight patch angle in Local Detail

Fig. 4. The key parameters of the generated path by RFPPO.

Fig. 4 shows the key parameters of the path generated by RFPPO in the task of Fig. 3 (e), when facing an obstacle moving in a straight line. Fig. 4 (a) is the vertical plane view of the flight path, where the horizontal axis represents the cumulative flight distance and the vertical axis represents the altitude, both in meters. The yellow line in the graph represents the aircraft's altitude, and the blue line represents the terrain's altitude. The graph shows that the aircraft's flight altitude contour consistently matches the terrain contour, demonstrating excellent terrain-following capabilities. Fig.4 (b) shows the ground clearance of the flight path. It is observed that the aircraft's ground clearance is stably maintained within the set range (450,550). Additionally, as our threat object appears near 20 kilometers, there is a noticeable smooth fluctuation in the aircraft's altitude for avoidance. Furthermore, Fig. 4 (c) and (d) illustrate the changes in the aircraft's climb angle and flight path angle during a 50 km mountainous flight, indicating how these angles vary at different flight distances. The pitch angle and flight path angle of the aircraft used in this study are set between -25 and 25 degrees. The graph reveals that the path planned by the algorithm fully meets the aircraft's dynamic constraints.

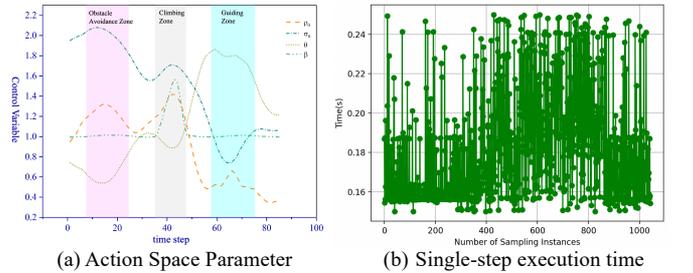

(a) Action Space Parameter  (b) Single-step execution time

Fig. 5. Action Space and Decision Time.

We analyzed a segment of the action space parameters from the task in Fig. 3 (a) to evaluate if our constructed field can rapidly respond to environmental changes. As shown in Fig. 8 (a), within the obstacle avoidance area, both the repulsion coefficient and tangential coefficient exhibit a rapid response. Upon entering the climbing area, the ground repulsion coefficient quickly increases. During this phase, as the distance between the agent and the obstacle gradually increases, the tangential field guiding the agent around the obstacle weakens. After entering the guidance area, a continuous change in the tangential angle is observed, guiding the agent to converge towards the target point and complete the flight objective under the influence of the tangential angle. Fig. 8 (b) displays the single decision time in our task, demonstrating that our algorithm can achieve path planning within 25ms even under low computational power.

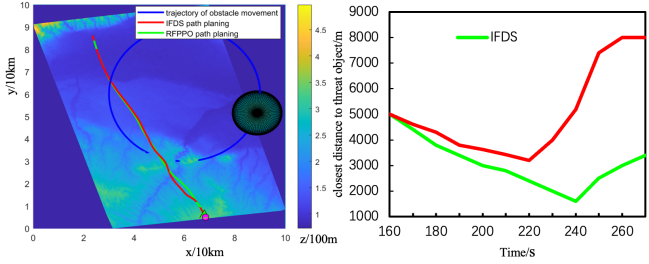

(a) overhead view  (b) Closest distances to the dynamic threat
Figure 1. Comparison between IFDS and RFPPO

To demonstrate the optimality of our algorithm, we compared RFPPO with IFDS and RRT* (which only performs global planning) algorithms under the same dynamic constraints, search step length, and dynamic limitations as our method. In the specific task requirements of this paper, we set the hyperparameters as follows: $\rho = 0.6$, $\sigma = 0.8$, and $\theta = 1.1$. Additionally, when the agent enters the mountainous area, a threat object performing sinusoidal movements is deployed 5KM away from the agent. Fig.6 (a) presents the overhead view of the entire process for both IFDS and RFPPO.

TABLE I.  FLIGHT PARAMETER COMPARISON CHART

| algorithm | Flight parameters | | |
|---|---|---|---|
| | path planning length | maximum climb angle | smoothness |
| RFPPO | 108 $Km$ | 15.19° | 0.1297 |
| IFDS[9] | 131 $Km$ | 25° | 0.3498 |
| RRT*[18] | 129 $Km$ | 15.58° | 0.2917 |

As shown in Fig.3 (b), during the process from simultaneously detecting a threat to moving away from it, the RFPPO algorithm ensures a significantly greater minimum distance between the aircraft and obstacles compared to the IFDS algorithm. This indicates that the RFPPO algorithm can respond to threats more swiftly. Additionally, as seen in Tab.1, RFPPO outperforms other algorithms in terms of the total flight distance, flight state, and trajectory smoothness throughout the flight process. The smoothness metric, defined as the sum of the squared angles at each segment of the path divided by the total number, is smaller for smoother trajectories, and RFPPO scores better on this measure.

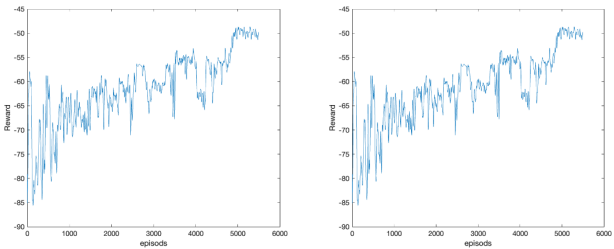

Figure 2. *Comparison between IFDS and RFPPO.*

Fig.8 illustrates how the model's convergence varies with training episodes. Specifically, during training, we conduct an experiment like in Fig.3 (a) after every 10 training sessions and calculate the overall return. Fig.8 (b) shows the results using the key point method based on RRT*. With this method, the model converges after just 3000 training iterations. In contrast, without this approach, convergence requires over 5000 training sessions, and the final overall return is comparatively lower.

This framework, without relying on prior global planning, achieves collision-free trajectory planning that conforms to dynamic constraints. It demonstrates the potential of combining proximal policy gradient algorithms with traditional path planning methods in addressing challenges traditional path planning faces, such as following complex three-dimensional terrain and adhering to kinematic constraints.

IV. CONCLUSION

For large and medium-sized fixed-wing aircraft operating in complex environments, this study proposes the RFPPO flight path planning algorithm based on the aircraft motion dynamics model. Tailoring to mission characteristics, we restructured the state and action spaces of reinforcement learning through APF and IFDS, while strictly adhering to aircraft dynamics in state transition. Additionally, we introduced a Key Point training strategy based on Motion Dynamics RRT, significantly reducing the computational resources required for training. Extensive simulations on DEM maps were conducted, developing an effective reward function that converges efficiently. The results demonstrate that the algorithm can effectively perform TF/TA in highly complex environments, achieving dynamic obstacle avoidance within 25 milliseconds. In the future, we look forward to further exploring the potential of Transformer networks in the flight state recognition of large and medium-sized aircraft and their application in disturbance flow field control for multi-aircraft collaborative avoidance.